\documentclass[letterpaper]{article} 
\usepackage[submission]{aaai25}  
\usepackage{times}  
\usepackage{helvet}  
\usepackage{courier}  
\usepackage[hyphens]{url}  
\usepackage{graphicx} 
\usepackage{natbib}  
\usepackage{caption} 
\usepackage{algorithm}
\usepackage{algpseudocode}
\usepackage{newfloat}
\usepackage{listings}
\DeclareCaptionStyle{ruled}{labelfont=normalfont,labelsep=colon,strut=off} 
\floatstyle{ruled}
\newfloat{listing}{tb}{lst}{}
\floatname{listing}{Listing}

\usepackage{xurl}
\usepackage{kotex}
\usepackage{tabularray}
\usepackage{newtxmath}
\usepackage{colortbl}
\usepackage{soul}
\usepackage{comment}
\usepackage{enumitem}
\usepackage{tcolorbox}

\usepackage{footnote}

\usepackage{multirow}
\usepackage{multicol}
\usepackage{arydshln}
\usepackage{subfig}
\usepackage{adjustbox}

\usepackage{amsthm}
\usepackage{amsmath}

\usepackage{listings,lstautogobble}
\newcommand{\review}[1]{{\color{blue}{#1}}} 
\newcommand{\fixme}[1]{{\color{red}{#1}}} 

\usepackage{dblfloatfix}
\usepackage{fixltx2e}
\newcommand{\cancel}[1]{{\color{red}{#1}}}

\usepackage{caption}

\definecolor{forestgreen}{rgb}{0.0, 0.27, 0.13}

\newcommand{\sj}[1]{{\color{blue}{#1}}}

\definecolor{darkblue}{rgb}{0.0, 0.0, 0.54}

\newcommand{\FIG}[1]{Fig.~\ref{#1}}

\newcommand{\TAB}[1]{Table~\ref{#1}}

\usepackage{tikz}

\newcommand{\wcircled}[1]{
    \setbox0=\hbox{#1}%
    \dimen0\wd0%
    \divide\dimen0 by 2%
    \begin{tikzpicture}[baseline=(a.base)]%
        \useasboundingbox (-\the\dimen0,0pt) rectangle (\the\dimen0,1pt);
        \node[color=black!50!black,circle,draw,outer sep=0pt,inner sep=0.05ex] (a) {#1};
    \end{tikzpicture}
}

\newcommand{\bcircled}[1]{
    \setbox0=\hbox{#1}%
    \dimen0\wd0%
    \divide\dimen0 by 2%
    \begin{tikzpicture}[baseline=(a.base)]%
        \useasboundingbox (-\the\dimen0,0pt) rectangle (\the\dimen0,1pt);
		\node[fill=black!100,circle,text=white,outer sep=0pt,inner sep=0.05ex] (a) {#1};
    \end{tikzpicture}
}

\newcommand{\redcircled}[1]{
    \setbox0=\hbox{#1}%
    \dimen0\wd0%
    \divide\dimen0 by 2%
    \begin{tikzpicture}[baseline=(a.base)]%
        \useasboundingbox (-\the\dimen0,0pt) rectangle (\the\dimen0,1pt);
		\node[fill=red!100,circle,text=white,outer sep=0pt,inner sep=0.05ex] (a) {#1};
    \end{tikzpicture}
}

\newcommand{\bluecircled}[1]{
    \setbox0=\hbox{#1}%
    \dimen0\wd0%
    \divide\dimen0 by 2%
    \begin{tikzpicture}[baseline=(a.base)]%
        \useasboundingbox (-\the\dimen0,0pt) rectangle (\the\dimen0,1pt);
		\node[fill=blue!100,circle,text=white,outer sep=0pt,inner sep=0.05ex] (a) {#1};
    \end{tikzpicture}
}

\usepackage{comment}

\newcommand{\ours}[0]{\textsc{SeDi-Instruct}}
\def\thickhline{\noalign{\hrule height1.1pt}}

\title{\ours{}: Enhancing Alignment of Language Models through Self-Directed Instruction Generation}
\author{
    Jungwoo Kim,
    Minsang Kim,
    Sungjin Lee
}
\affiliations {
    Daegu Gyeongbuk Institute of Science and Technology\\
    jungwoo@dgist.ac.kr, kimmsang96@dgist.ac.kr, sungjin.lee@dgist.ac.kr
}

\usepackage{bibentry}

\begin{document}

\maketitle

\begin{abstract}
The rapid evolution of Large Language Models (LLMs) has enabled the industry to develop various AI-based services. 
Instruction tuning is considered essential in adapting foundation models for target domains to provide high-quality services to customers.
A key challenge in instruction tuning is obtaining high-quality
instruction data. Self-Instruct, which automatically 
generates instruction data using ChatGPT APIs, 
alleviates the data scarcity problem. To improve the
quality of instruction data, Self-Instruct discards 
many of the instructions generated from ChatGPT, 
even though it is inefficient in terms of cost owing 
to many useless API calls.
To generate high-quality instruction data at a low cost,
we propose a novel data generation framework, \textbf{Se}lf-\textbf{Di}rect
\textbf{Instruct}ion generation (\ours{}), which employs diversity-based filtering and iterative feedback task generation.
Diversity-based filtering maintains model accuracy 
without excessively discarding low-quality generated instructions 
by enhancing the diversity of instructions in a batch. 
This reduces the cost of synthesizing instruction data.
The iterative feedback task generation integrates
instruction generation and training tasks and utilizes information
obtained during the training to create high-quality instruction
sets.
Our results show that \ours{} enhances the accuracy 
of AI models by 5.2\%, compared with traditional methods, 
while reducing data generation costs by 36\%.    
\end{abstract}

%

\section{Introduction}
\label{sec:intro}


Many novel Artificial Intelligence (AI)-based services have been emerging in the
wake of the Large Language Models (LLMs). ChatGPT, a representative LLM-based
service, recorded approximately 1.6 billion visitors worldwide in December 2023
and was reported to be used by more than 100 million people
weekly~\cite{chatgpt, chatgptuser}. Additionally, many companies have already
introduced AI or are considering introducing it. This suggests that AI is going
beyond mere technological innovation and is fundamentally changing various
industries.

To effectively leverage AI technologies in the industry, a substantial amount of
domain-specific data is essentially required, 
as the performance of AI models heavily
depends on the quantity and quality of data.
However, it is challenging to obtain data that reflects diverse
real-world scenarios and that is tailored to specific domains. 
Additionally, security, privacy, and ethics issues further 
complicate data collection efforts. The average
cost of assembling a dataset is prohibitively high.
For example, the price of high-quality language
datasets is up to \$0.15/word~\cite{datasetprice}.

To address the challenges of data collection, various
techniques such as data augmentation~\cite{backtranslation,eda,self-instruct}, data-efficient
training~\cite{data-efficient}, and transfer learning~\cite{torrey2010transfer} have been proposed.
Among these, Self-Instruct, which automatically generates 
large amounts of datasets, has received significant attention 
for its ability to mitigate data shortages.
Self-Instruct operates in the following three steps:
The first step is a generation phase. 
By supplying \textit{seed instructions} prepared by humans
to LLM services like ChatGPT,
it automatically synthesizes a set of instructions,
which are called \textit{generated instructions}.
The second step is a filtering phase, which
eliminates duplicates and useless ones 
(that negatively impact training efficiency)
from the generated instructions.
Through this step, only meaningful instructions, 
called \textit{kept instructions}, are selectively chosen and 
kept for future use.
The final step is a training phase, where AI models are trained
using the kept instructions. 
Self-Instruct can mitigate the difficulties associated with
domain-specific and large-scale instruction data collection by automating data generation.

Despite its potential, Self-Instruct does have limitations
that need to be addressed.
One significant limitation of Self-Instruct is 
\textit{low data quality}.
A substantial portion of the generated data (or instructions) is inaccurate,
compromising the overall effectiveness of instruction tuning. 
Prior studies have shown that
a model trained using only 10\% of the kept instructions
achieves similar or even better accuracy
than one trained using the entire kept instructions~\cite{alpagasus}.
This indicates that a considerable amount of the kept instructions is 
inaccurate and is not effective in training models,
highlighting the need for improved
data generation and validation processes to enhance the
quality of the dataset.

\begin{figure}[t]
	\centering
	\subfloat[Ratio of kept instructions to API requests]{\includegraphics[width=0.428\linewidth]{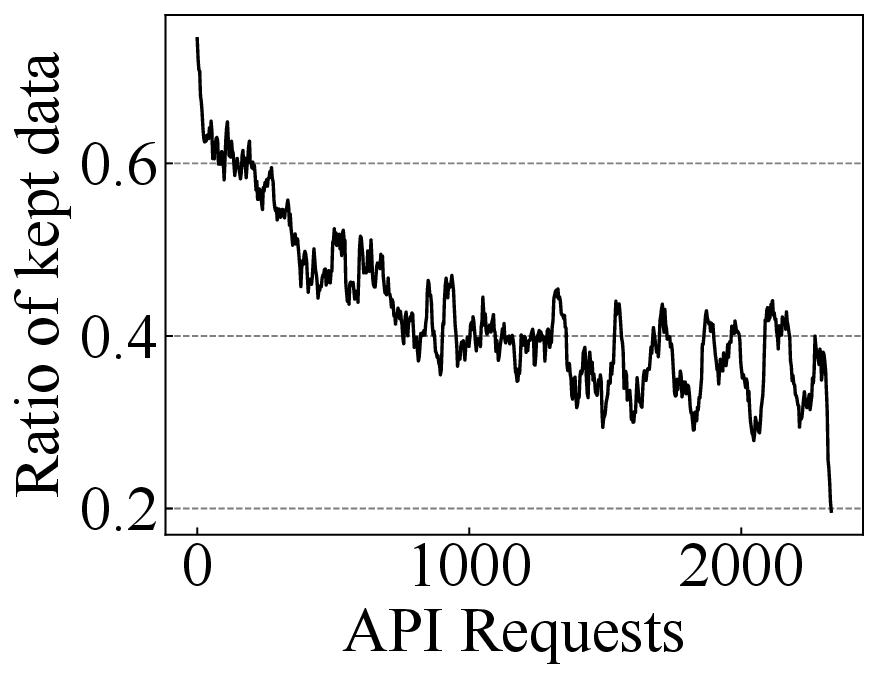}}
	\hspace{3pt}
	\subfloat[Comparison of \# of generated instructions and kept]{\includegraphics[width=0.442\linewidth]{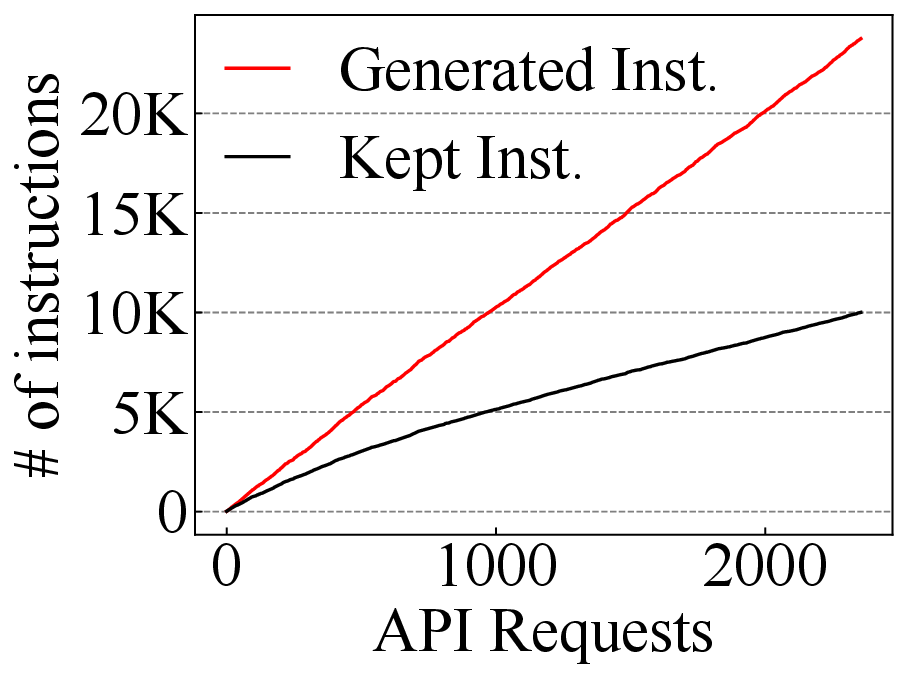}}
	\caption{Filtering inefficiency problem}%
	\label{fig:api-inefficiency}%
\end{figure}

Another limitation of Self-Instruct is \textit{filtering inefficiency}. 
To improve the quality of kept instructions used for training,
Self-Instruct discards a significant percentage of generated instructions
by performing the filtering process. This inevitably increases the number of 
useless ChatGPT API calls that are actually not needed.
\FIG{fig:api-inefficiency}(a) illustrates the ratio of kept instructions 
to the number of ChatGPT API requests we actually made over time.
\FIG{fig:api-inefficiency}(b) presents the accumulated numbers of kept and 
generated instructions for \FIG{fig:api-inefficiency}(a).
To create 10K kept instructions,
Self-Instruct needs to generate 24K
instructions, meaning that approximately 58\% of the generated data
is abandoned~\cite{self-instruct}.
This trend becomes increasingly severe as the amount of data grows,
showing inefficiencies in the filtering process.
This implies that the excessive number of ChatGPT API invocations
is unnecessary, 
resulting in a waste of computing resources in data centers 
and higher operational expenses.
Consequently, more efficient filtering mechanisms are highly needed.

\begin{figure}
    \centering
    \includegraphics[width=0.75\linewidth]{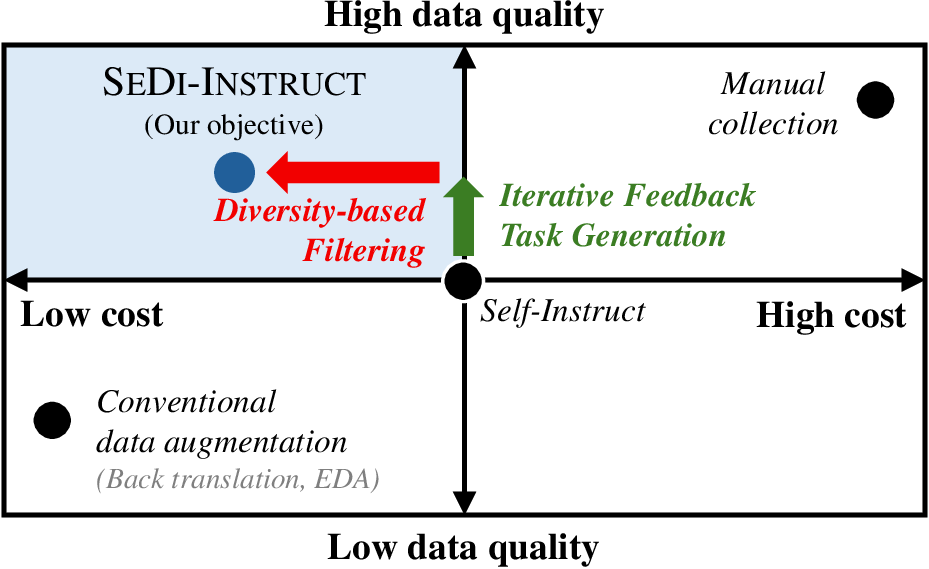}
    \caption{Goal of \ours{}}
    \label{fig:objective}
\end{figure}

In this paper, we propose a novel data generation
framework, \textbf{Se}lf-\textbf{Di}rect
\textbf{Instruct}ion generation, \ours{} in short.
As illustrated in \FIG{fig:objective}, the ultimate goal of \ours{}
is to generate high-quality instructions at low costs, beating other 
data generation approaches~\cite{self-instruct,sitgpt4,alpaca}
that require high data collection costs (e.g., manual collection~\cite{llama3}) or 
produce low-quality data (e.g., data augmentation~\cite{backtranslation, eda}).
\ours{} achieves this by effectively addressing 
the issues of low data quality and filtering inefficiency.

The design of \ours{} is based on two insights derived from prior literature.
First, \ours{} tackles the low data quality problem
by leveraging the inherent properties of few-shot learning, 
which Self-Instruct is based on. 
We focus on the fact that the performance
of the few-shot learning is
directly influenced by the quality of the source data,
which corresponds to the seed instructions in Self-Instruct.
\ours{} improves the quality of seed instructions by employing
an \textit{iterative feedback task generation technique}
that integrates the training and generation processes.
During training, \ours{} extracts measures indicating the quality of 
the seed data and uses these to refine and update existing seed
instructions without any human involvement.
In this way, \ours{} continually provides high-quality seed data
for instruction tuning.

Second, we address the filtering inefficiency problem
by balancing the distribution
of kept instructions over batches.
Previous studies reported that even if
a training dataset is relatively skewed (that is,
the label or task distribution of the training dataset is imbalanced),
we can minimize the negative impact by making each batch 
have balanced data~\cite{8215530}.
Keeping this property in mind, \ours{} employs a \textit{diversity-based filtering
technique}. \ours{} generates instructions in the same manner
as Self-Instruct but generously accepts (low-quality) instructions 
that are moderately similar to previously kept ones.
Instead, improving the diversity of instructions 
within a batch can maintain comparable accuracy
without discarding many of the generated instructions. As a result,
it lowers the cost of generating instructions.

To evaluate \ours{}, we train two models using Llama-3-8B:
one utilizing \ours{} and the other leveraging the Self-Instruct.
To benchmark our models against the ideal scenario where Llama-3-8B
operates at its maximum potential,
we compare \ours{} with the Llama-3-8B-Instruct model that is
tuned with 10M human-written instructions.
We also include the Falcon-7B-Instruct and Gemma-7B-Instruct
models, whose parameter sizes are similar to \ours{}.
Our results show that the model trained with \ours{} not only outperforms
the existing models except for Llama-3-8B-Instruct 
but also achieves significant cost efficiency, reducing expenses by up to 36\% compared to Self-Instruct.

\section{Related Work}
\label{sec:relatedwork}

\begin{figure}[b]
    \centering
    \includegraphics[width=0.82\linewidth]{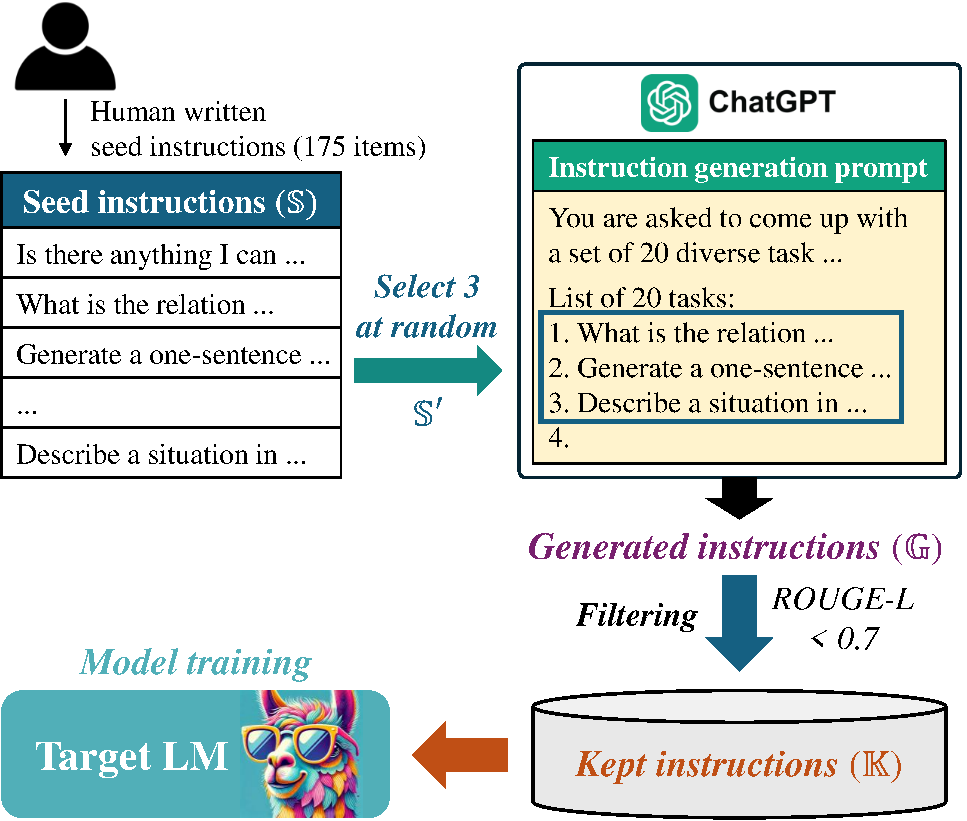}
    \caption{Overall organization and operations of Self-Instruct}
    \label{fig:self-instruct}
\end{figure}

\begin{figure*}[t!]%
	\centering
	\includegraphics[width=0.9\linewidth]{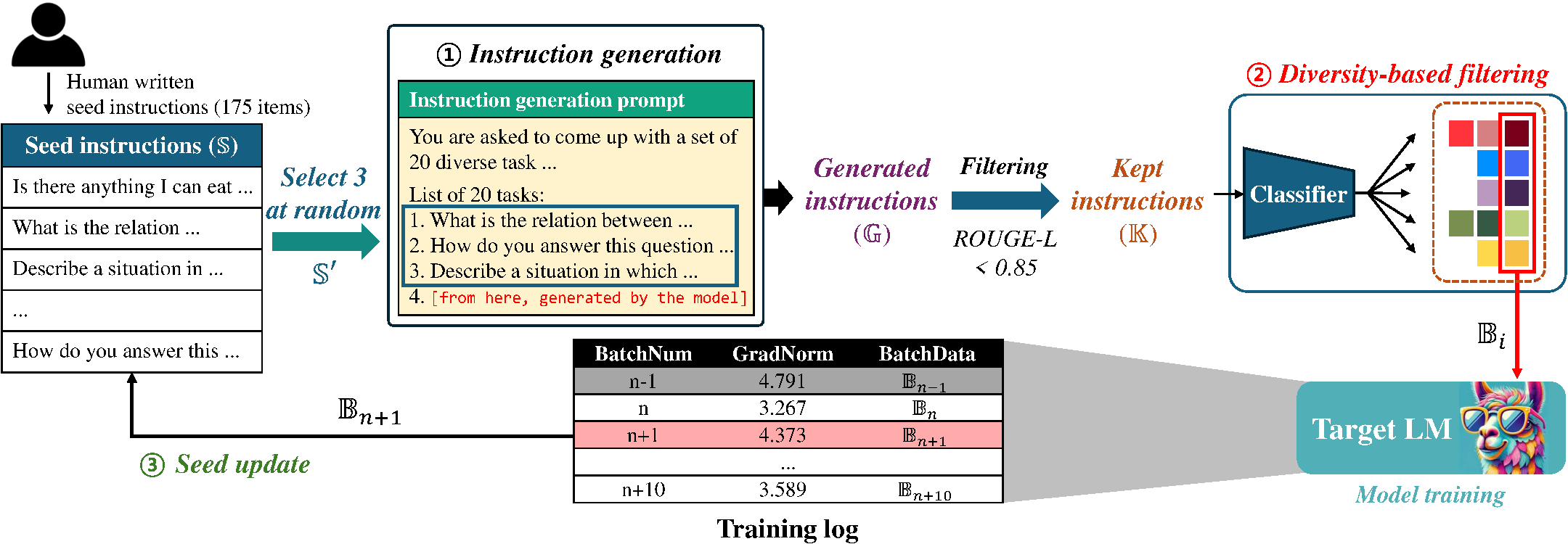}
	\caption{Overall organization and operations of \ours{}}%
	\label{fig:sedi-instruct}%
\end{figure*}
\subsection{Instruction Tuning}
Instruction tuning aims to enhance the performance of pretrained LLMs by
fine-tuning. Typically, a pretrained LLM such as GPT-3~\cite{gpt3},
PaLM~\cite{PaLM}, and LLaMA-3~\cite{llama3} generates outputs based
on statistical patterns learned from vast amounts of text data, predicting the
next word in a sequence. However, these predictions often differ from user
expectations because the model selects the highest-probability token rather than
generating a response tailored to the specific request.

Instruction tuning addresses this issue by aligning model outputs more closely
with human expectations. Specifically, to ensure that the model works as
expected, 
fine-tuning is conducted using 
an instruction dataset that consists of input queries and desired responses.
Notably, instruction-tuned models like FLAN-PaLM~\cite{FLAM},
InstructGPT~\cite{InstructGPT}, and gemma-it~\cite{gemmait} have significantly
improved zero-shot performance. This indicates that instruction tuning enables
models to better leverage the existing knowledge embedded within their parameters
from pretraining.

Despite its potential, instruction tuning encounters significant challenges,
particularly related to data scarcity. Effective instruction tuning demands
large, diverse, and high-quality instruction datasets to ensure learning across
various contexts. However, the availability of such datasets is often limited,
impeding the model's ability to generalize and perform well on unseen tasks.

\subsection{Overcoming Challenges of Limited Datasets}
One effective solution for overcoming data scarcity is data augmentation.
Back translation~\cite{backtranslation} translates text data into
another language and then back to the original language, generating textual data
with different words while preserving the original context. Jason Wei and Kai
Zou proposed a straightforward data augmentation technique called
Easy Data Augmentation (EDA), which consists of four simple operations: synonym
replacement, random insertion, random swap, and random deletion. These
techniques help augment datasets, improving model performance~\cite{eda}. However,
replicating the dataset may not sufficiently improve task and context diversity.

One fundamental solution to the data collection challenge is leveraging
LLMs for data synthesis. 
The representative framework is Self-Instruct~\cite{self-instruct}. 
As shown in \FIG{fig:self-instruct},
Self-Instruct utilizes prompt engineering
and produces generated instructions ($\mathbb{G}$) by feeding
a randomly chosen subset ($\mathbb{S}'$) of seed instructions ($\mathbb{S}$) 
to ChatGPT API.
The generated instructions are then subjected
to a filtering process that compares the ROUGE-L similarity
with kept instructions ($\mathbb{K}$).
If the similarity between a newly generated
instruction and kept instructions exceeds
a specified threshold (indicating
that the new instruction is 
similar to those already in the dataset),
the instruction will be excluded from the training dataset.
Otherwise, it is included in $\mathbb{K}$ as a new unique instruction.
This approach 
alleviates data scarcity 
and improves
the diversity of the synthesized datasets, making it a promising strategy for
training robust AI models.

However, the instruction dataset synthesized by the Self-Instruct,
such as the ALPACA 52K dataset~\cite{alpaca}, often contains a significant amount of 
inaccurate or irrelevant responses, leading to low-quality data.
This is due to the lack of a validation or feedback process beyond
simple similarity-based filtering. 
As a result, there is a strong need 
for a post-validation process to improve the quality of instructions.

Alpagasus~\cite{alpagasus} addresses this issue 
by using LLM filters (e.g., ChatGPT)
to further filter out the kept instructions of the ALPACA 52K dataset,
creating a refined dataset of 9K high-quality entries. 
Remarkably, this smaller and high-quality
dataset achieves performance equal to or better than the original 52K dataset.
Additionally, Feng et al.~\cite{modelcollapse1} demonstrated that by
filtering out 87.5\% of a synthesized dataset using human verifiers, the resulting
model outperforms one trained on the entire dataset.

\section{Method}
\label{sec:method}

In this section, we present our data generation framework,
\ours{}.
\FIG{fig:sedi-instruct} illustrates 
the overall organization and operations of \ours{}.
In contrast to Self-Instruct shown in~\FIG{fig:self-instruct},
\ours{} combines the training and generation processes in a pipeline manner
so that the two components interplay to produce better outputs at a lower cost.
Similar to Self-Instruct, \ours{} uses LLMs to generate the 
instructions $\mathbb{G}$ 
using a subset of seed instructions $\mathbb{S}'$ 
that are randomly chosen 
from the entire seed instruction set $\mathbb{S}$
(see \wcircled{1}). 
Diversity-based filtering then composes
a set of kept instructions $\mathbb{K}$
by filtering out useless ones in $\mathbb{G}$ (\wcircled{2}). 
Finally, through the iterative feedback task generation,
\ours{} replaces some seed instructions in $\mathbb{S}$
with new ones in $\mathbb{B}_{n+1}$ which are selected by analyzing 
training logs collected during the training process (\wcircled{3}).

In the rest of this section, we describe the diversity-based filtering 
and then present how the iterative feedback task generation operates.

\subsection{Diversity-based Filtering}

Many approaches to enhancing the effectiveness of Self-Instruct have focused on
retaining high-quality instructions~\cite{alpagasus, modelcollapse1}. 
To achieve this, they attempt to eliminate inaccurate and 
redundant instructions through simple heuristics, 
such as ranking the instructions and removing those with low ranks. 
While this is effective, excessive filtering results in many
unnecessary inferences that involve inefficient API usage and resource waste. 

To mitigate this inefficiency, our diversity-based filtering aims to minimize
discarded instructions by allowing slight redundancy in the kept instructions.
The existing filtering method uses a ROUGE-L similarity threshold of 0.7, 
which is relatively tight,
but we loosen it 
to 0.85. With the threshold of 0.7, 
a moderate number of redundant instructions, 
accounting for about half of the generated instructions, are removed.
In our case, with the threshold of 0.85, 
only instructions that exhibit significant redundancy are discarded, 
which accounts for roughly 20\% of the total instructions.

This strategy, however, poses a risk of reducing diversity, potentially leading
to a skewed distribution in the overall dataset. 
This decline in global diversity can degrade 
training efficiency.
We address this problem by enhancing the local diversity within
each batch.
This strategy minimizes the
negative impact of data redundancy on the model, as it helps the loss function
to learn uniformly, thereby improving the stability of the training process.
Consequently, it contributes to better generalization performance, even when
the data distribution is a little biased~\cite{batchdiversity2, batchdiversity3}.

To maximize diversity within each batch, 
we classify the generated instructions into clusters based on their similarity. 
The number of clusters matches the batch size. 
For example, with a batch size of 16, we create 16 clusters, 
each containing similar instructions. When forming a batch, we select one instruction from each cluster, one by one, and include it in the batch.

\begin{algorithm}[t]
\begin{algorithmic}[1]
\caption{Diversity-based filtering}
\label{alg:diversity-filtering}
\footnotesize
\Statex \hspace{-\algorithmicindent} \textbf{Input:} ROUGE-L similarity threshold ($\theta_{ROUGE}$), seed instructions ($\mathbb{S}$), kept instructions ($\mathbb{K}$), Clusters ($\mathbb{C}$)

\Statex \hspace{-\algorithmicindent} \textbf{Output:} Clusters ($\mathbb{C}$)
\vspace{-5pt}
\Statex \hspace{-\algorithmicindent} \rule{230pt}{.2pt}

\State $\mathbb{S}' \gets \text{random.sample}(\mathbb{S},3)$
\State $\mathbb{G} \gets \textsc{GenerateInstruction}(\mathbb{S}')$
\For{\textbf{all} $gen_{i} \in \mathbb{G}$}
    \State $isVariant \gets \text{True}$
    \For{\textbf{all} $kept_{j} \in \mathbb{K}$}
        \State $sim \gets \textsc{ROUGE-L-Similarity}(gen_{i}, kept_{j})$
        \If{$sim > \theta_{ROUGE}$}
            \State $isVariant \gets \text{False}$
            \State \textbf{break}
        \EndIf
    \EndFor
    \If{$isVariant$}
        \State $\mathbb{K}.\text{add}(gen_{i})$
        \State $class \gets \textsc{Classifier}(gen_{i})$
        \State $\mathbb{C}[class].\text{put}(gen_{i})$
    \EndIf
\EndFor
\State \Return $\mathbb{C}$
\end{algorithmic}
\end{algorithm}

\textbf{Filtering Algorithm}
Algorithm \ref{alg:diversity-filtering} illustrates our filtering algorithm.
The initial step involves extracting $\mathbb{S}'$, a subset of
$\mathbb{S}$, by randomly selecting three elements from $\mathbb{S}$ (Line 1).
Next, it generates a new set of generated
instructions $\mathbb{G}$ based on $\mathbb{S}'$ using a language model
such as ChatGPT (Line 2).
For each generated instruction $gen_i$ in $\mathbb{G}$, 
the algorithm computes its ROUGE-L similarity score $sim$ 
with each instruction $kept_j$ in a kept instruction set $\mathbb{K}$ (Lines 3--6). 
Note that $\mathbb{K}$ contains unique instructions included in the final 
instruction pool.
If there exists at least one $kept_j$ instruction whose
$sim$ exceeds the threshold $\theta_{ROUGE}$,
$gen_i$ is treated as redundant, and $isVariant$ is set to False
to prevent its inclusion in the final pool (Lines 7--11).
On the other hand, if $sim$ is below $\theta_{ROUGE}$ for all $kept_i$ in $\mathbb{K}$,
$gen_i$ is added to $\mathbb{K}$, 
indicating that the instruction $gen_i$ is included 
into $\mathbb{K}$ as a new unique instruction (Lines 12--13).

As we explained before, \ours{} uses the relaxed similarly threshold,
$\theta_{ROUGE} = 0.85$, compared to the existing designs.
To minimize the negative impact of the reduced diversity on model training, 
we employ clustering-based batch configure method. 
Once the instruction $gen_i$ is accepted to be included in $\mathbb{K}$,
it is classified into an appropriate category $class$ using 
a predefined classifier model \textsc{Classifier} (Line 14). 
This classifier vectorizes the instructions and performs  
Principal Component Analysis (PCA) to reduce the dimensionality 
to $\log({batch\_size})$. The
resulting dimensions from PCA are then used to perform clustering based on
the quadrants in the reduced space. 
This approach produces a number of clusters equal to the batch size.
This categorization allows the
algorithm to group instructions into clusters $\mathbb{C}$, 
which are then used to construct batches for training (Line 15).

During the training, 
an instruction is popped from each cluster to form a batch
when the \texttt{\_\_getitem\_\_} function that creates a batch 
to feed to the model is called.
Through this, each batch is formed to encompass a diverse range of
instructions, promoting the model's ability to generalize across various
contexts and minimizing the need for strict filtering. Therefore, by employing a
relaxed ROUGE-L similarity threshold and clustering similar instructions 
in a batch, the algorithm reduces the unnecessary elimination of instructions 
and maximizes the retention of helpful information.

\begin{figure}[t]
    \centering
    \includegraphics[width=0.95\linewidth]{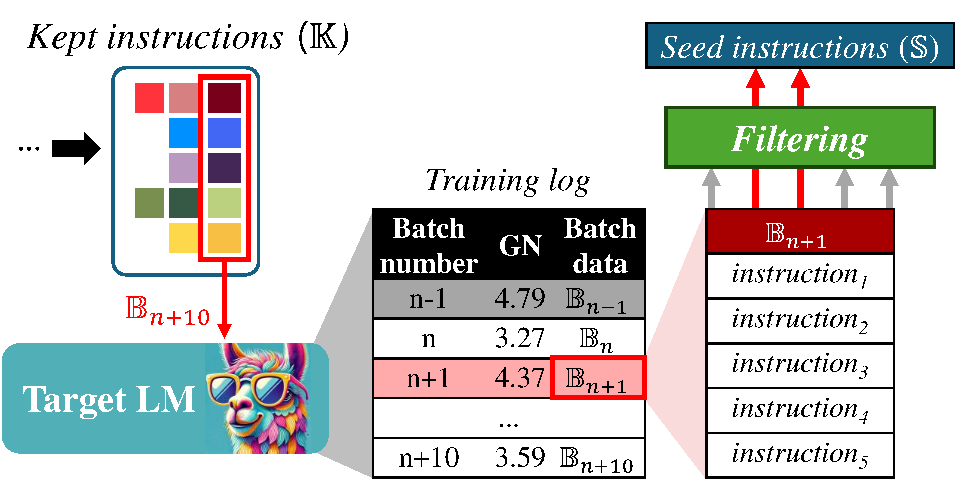}
    \caption{Identification of attractive batches and instructions}
    \label{fig:seed-update}
\end{figure}

\subsection{Iterative Feedback Task Generation}

As we explained in Introduction,
the quality of \ours{} is highly affected by the quality of seed instructions.
\ours{} aims to make the quality of seed instructions better 
by adding new seed instructions to $\mathbb{S}$ if they can lead to better model training and
evicting ones from $\mathbb{S}$ if they are less effective for the training.
\ours{} first finds good candidate batches and instructions 
by leveraging insights gained from the training phase.
Then, it identifies valueless seed instructions in the set $\mathbb{S}$.
Finally, it adds new seed instructions to $\mathbb{S}$, while
removing valueless ones.

\textbf{Finding Candidate Instructions}
\FIG{fig:seed-update} illustrates how \ours{} identifies attractive
batches and chooses high-quality instructions that will
be added to the seed instruction set $\mathbb{S}$.
During training, \ours{} records information about batches $\mathbb{B}_i$
in \textit{Training log}.
The information in the log reflects the training quality and is subsequently used 
to select a high-quality batch. 
There exist various metrics to measure the quality of batches for training,
which may include loss variance, gradient norm, and the sum of both.
Based on our empirical study, we decide to use the gradient norm (GN) because 
it is the most suitable one, reflecting the training quality of batches.

For every ten iterations, we identify the batch with the highest gradient norm. 
For the first ten iterations, we do not choose any batches 
as the gradient norms at these iterations 
do not accurately reflect the quality of training.
Once a batch $\mathbb{B}_{i}$ is selected, we look for candidate seed instructions
in the batch.
In the same way as Lines 3--13 of Algorithm~\ref{alg:diversity-filtering}, 
we choose instructions in $\mathbb{B}_{i}$ that are not similar to those in $\mathbb{S}$. More specifically,
we only include instructions whose ROUGE-L similarity $sim$ exceeds 
$\theta_{ROUGE}=0.7$. $\theta_{ROUGE}$ of 0.7 is tighter 
than 0.85 we used
for the diversity-based filtering. 
This is because every instruction in a batch already went through
the filtering process and thus always has
a lower ROUGE-L similarity than 0.85.
\begin{figure}[t]
    \centering
    \includegraphics[width=0.95\linewidth]{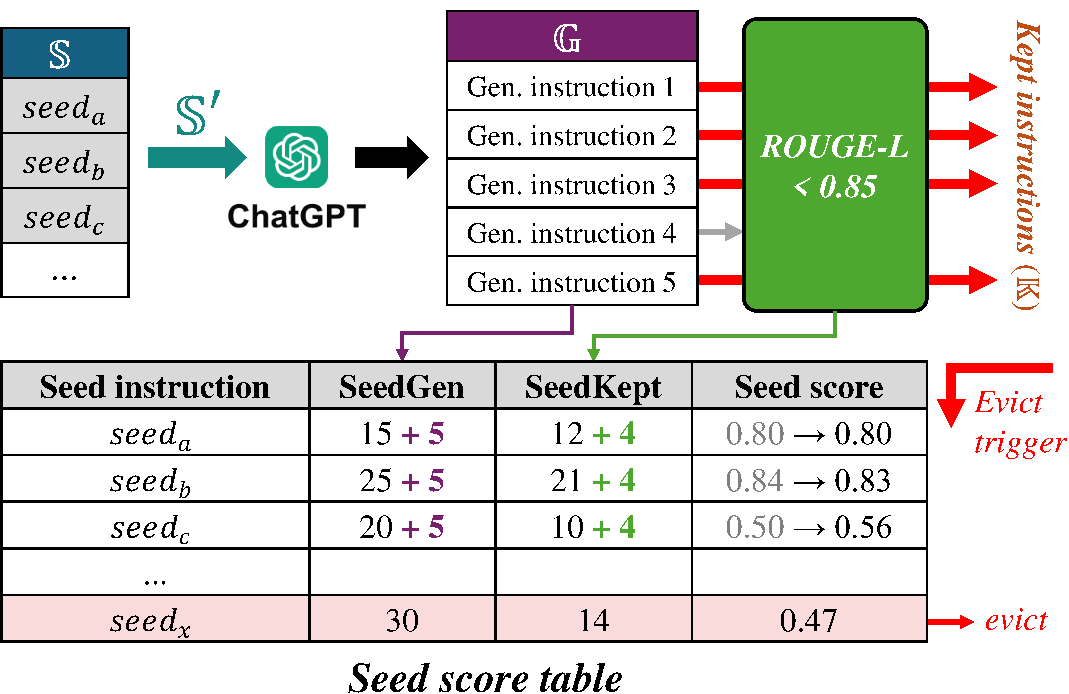}
    \caption{Selection of victim seed instructions in $\mathbb{S}$}
    \label{fig:seed-evict}
\end{figure}

\textbf{Finding Victim Seed Instructions}
Once candidate instructions to add are selected, we must choose 
the seed instructions to be removed from the set $\mathbb{S}$.
For the victim selection, we maintain a \textit{seed score}
for each seed instruction in a \textit{seed score table}. 
The seed score represents the diversity (or quality) of 
instructions generated from a particular seed instruction. 
This score is determined when we calculate the ROUGE-L similarity of 
generated instructions to decide whether or not to add them to 
$\mathbb{K}$.
If many generated instructions from a seed instruction are retained, 
it means that the seed instruction is of high quality.

\FIG{fig:seed-evict} illustrates an example of how \ours{} calculates
scores for seed instructions. To compute scores, 
the seed score table maintains two additional fields: 
SeedGen and SeedKept.
During the generation phase, a subset $\mathbb{S}'$ 
is selected from $\mathbb{S}$ in the same manner as we described
in Algorithm~\ref{alg:diversity-filtering} (see Line 1). 
Let us assume that seed instructions $seed_a$, $seed_b$, and $seed_c$ are
selected for $\mathbb{S}'$. 
Using them, \ours{} creates generated instructions $\mathbb{G}$. 
The number of instructions generated is randomly determined by the LLM;
in our example, five instructions are created.
The corresponding SeedGen field 
in the seed score table is then increased by 5.
Subsequently, in the filtering phase, 
one generated instruction is discarded through the filtering, and four survive,
which are then added to the kept instructions $\mathbb{K}$. 
Similarly, the corresponding SeedKept field in the table 
is increased by 4.
Finally, the scores of the three seed instructions are updated 
as the ratio of \textit{SeedKept} to \textit{SeedGen}. For example, the score of $seed_b$ was $21/25 = 0.84$, but it is updated to $(21+4)/(25+5) = 0.83$.

\textbf{Seed Replacement}
Seed instructions with low scores are evicted from $\mathbb{S}$
and are removed from the seed score table. 
Instead, more valuable instructions chosen from batches are added to $\mathbb{S}$.
This replacement of seed instructions is reasonable 
because a low score indicates that the seed instruction is unlikely to
generate diverse instructions.
In this manner, the iterative feedback task generation 
is able to keep qualified seed instructions for training.

\begin{table*}[t]
\caption{Summary of model accuracies over various benchmarks}
\label{tab:benchmarks}
\footnotesize
\centering
\begin{tabular}{l|cccc|c}
                            & \textbf{AlpacaEval}      & \textbf{MMLU (5-shot)} & \textbf{Hellaswag (0-shot)} & \textbf{ARC (0-shot)}   & \textbf{Average} \\ \cline{2-6} 
\textbf{Model}              & \textbf{Win \% vs GPT-4} & \textbf{Accuracy (\%)} & \textbf{Accuracy (\%)}      & \textbf{Accuracy (\%)}  & \textbf{}        \\ \thickhline
Llama-3-8B-Instruct         & 9.1                      & 65.7                   & 57.7                        & 72.2                    & 51.2             \\
Llama-3-8B + Self-Instruct  & 4.6                      & 56.5                   & 55.7                        & 67.7                    & 46.1             \\
\rowcolor[HTML]{DFDFDF} 
Llama-3-8B + \ours{} (ours) & 5.4                      & 56.6                   & 56.1                        & 69.3                    & 46.9             \\
Falcon-7B-Instruct          & 1.8                      & 25.1                   & 51.7                        & 62.0                    & 35.2             \\
Gemma-7B-Instruct           & 0.2                      & 50.2                   & 55.9                        & 66.3                    & 43.2             
\end{tabular}
\end{table*}

\section{Evaluation}
\label{sec:eval}

We conduct experiments to evaluate the effectiveness of \ours{}.
We first compare the performance of models trained with \ours{}
against other LLMs of a similar scale across various benchmarks. We then
carry out an in-depth analysis to ensure 
instruction tuning is executed correctly through
competitive evaluation. We also investigate 
whether the instruction generation of \ours{} is more cost-effective 
than that of Self-Instruct, mainly focusing on
how much diversity-based filtering reduces costs. 
Finally, we explore the impact
of model collapse and the potential safety issue of \ours{}.
In Appendix, a more detailed
investigation of the impact of the iterative feedback task generation, along
with other experiment details, are provided. 
All codes are available at \url{https://github.com/}. 

\subsection{Training Recipe}
\textbf{Our model.}
We use the Llama-3-8B model~\cite{llama3} as the base model
and train it with 30,164 instructions generated using \ours{}.
The seed instructions used at
the beginning of data generation are the same as those from
Self-Instruct. 
For detailed hyperparameters, please refer to Appendix.

\noindent \textbf{Baseline models.}
We compare \ours{} with four different models:
Llama-3-8B-Instruct, Llama-3-8B + Self-Instruct,
Falcon-7B-Instruct, and Gemma-7B-Instruct.
Llama-3-8B-Instruct represents the ideal instruction-tuned 
model. It is based on Llama-3-8B and is tuned using 10M 
manually collected instructions.
Llama-3-8B-Instruct has also been trained 
using Reinforcement Learning with Human Feedback (RLHF)~\cite{rlhf} 
and supervised fine-tuning (SFT)~\cite{sft1} to further 
enhance its performance. Such optimizations
enable Llama-3-8B-Instruct to outperform the other models.

Llama-3-8B + Self-Instruct is also based on Llama-3-8B,
but unlike Llama-3-8B-Instruct,
it is trained with instructions synthesized using Self-Instruct.

We also include Falcon-7B-Instruct and Gemma-7B-Instruct,
which have similar model sizes (7-8 billion parameters),
to evaluate \ours{} against models other than Llama-3-8B.
For detailed information on the models, please refer to Appendix.

\noindent \textbf{Hardware setup.} 
We use a machine that has two AMD EPYC 7742 3.3GHz
64-core CPUs and 2TB DDR4 DRAM. The machine is also equipped with eight RTX-3090
GPUs.
We use Ubuntu 22.04 as the OS and the version of
Python packages are torch 2.1.2 and deepspeed 0.14.4.

\subsection{Benchmark Performance}
We evaluate the models using
various datasets, including AlpacaEval~\cite{alpacaeval}, MMLU~\cite{MMLU}, Hellaswag~\cite{hellaswag},
and ARC~\cite{DROP}. 
We measure a win rate for AlpacaEval by comparing 
outputs from the models against those from GPT-4. 
A higher win rate indicates better alignment with expected responses. 
For the other benchmarks, we measure accuracy, 
representing the probability of correctly answering questions.
We measure the accuracy of MMLU in a 5-shot setting and 
the accuracy of Hellaswag and ARC in a zero-shot setting~\cite{scalingllm}.

\TAB{tab:benchmarks} shows the results,
where a higher value indicates better performance. 
Except for Llama-3-8B-Instruct which presents the ideal performance
with instruction tuning yet requires serious human efforts to create
seed instructions,
\ours{} outperforms all the other models we chose to compare.
Notably, despite using a more cost-effective data generation method,
\ours{} outperforms the Self-Instruct based model, showing 
5.2\% higher
accuracy on average. 
As will be discussed later, this higher accuracy is achieved with 36\%
lower training cost compared to Self-Instruct.

\begin{figure}[t]
    \centering
    \includegraphics[width=0.88\linewidth]{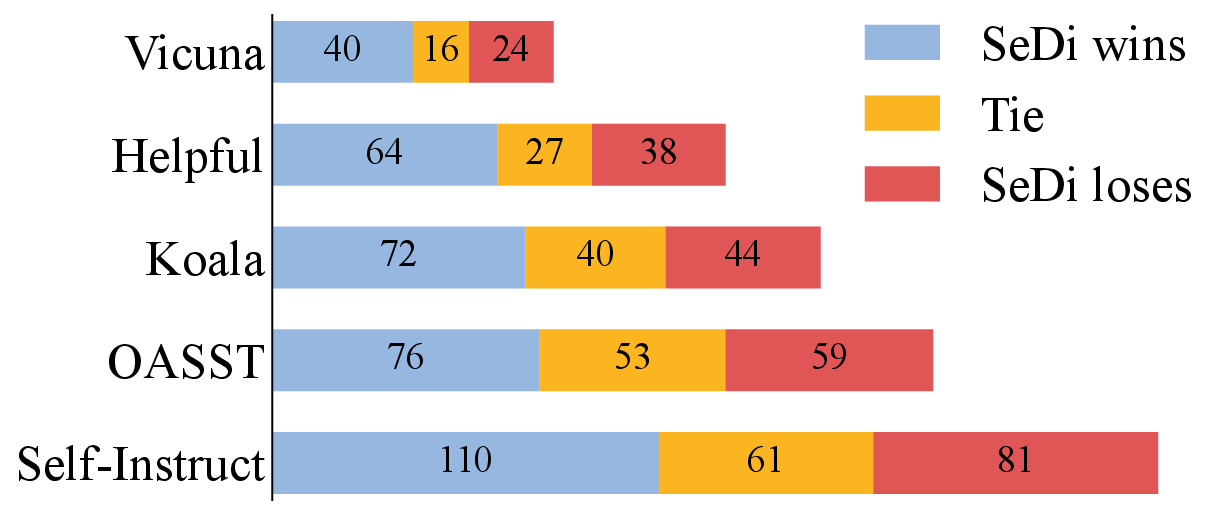}
    \caption{Competitive evaluation results}
    \label{fig:competitive}
\end{figure}

\subsection{Competitive Evaluation}
To assess the quality of the model's responses,
we make use of an automated competitive evaluation method 
that utilizes LLMs to compare the quality of responses~\cite{alpagasus, alpacafarm}. 
We compare our model (Llama-3-8B + \ours{}) 
with Llama-3-8B + Self-Instruct.
The responses from the models are input to GPT-4
which assigns a score between 1 and 10 for each response.
To mitigate a positional bias, 
we measure scores in two different orders:
first, when the responses from Llama-3-8B + \ours{} are input into GPT-4 
before those from Llama-3-8B + Self-Instruct, and second, when they are input afterward.
The final outcome is defined as ``Win-Tie-Lose"; ``Win" means our model wins twice for both orders, 
``Tie" means wins and loses once, and ``Lose" means our model loses twice.
The datasets used for the competition are the Vicuna test set (Vicuna)~\cite{vicuna1}, Anthropic's 
helpful test set (Helpful)~\cite{helpful}, the Koala test set (Koala)~\cite{koala}, the Open Assistant test set (OASST)~\cite{oasst}, 
and the Self-Instruct test set (Self-Instruct)~\cite{self-instruct}.

\FIG{fig:competitive} illustrates the results. 
As the results indicate, our model outperforms the 
Self-Instruct-based model for all of the five test sets.
It demonstrates that \ours{} generates more 
effective instructions for training than Self-Instruct through the iterative
feedback task generation.

\subsection{Cost Analysis}

\begin{figure}[t]%
	\centering
	\subfloat[\# of kept per API calls]{\includegraphics[width=0.55\linewidth]{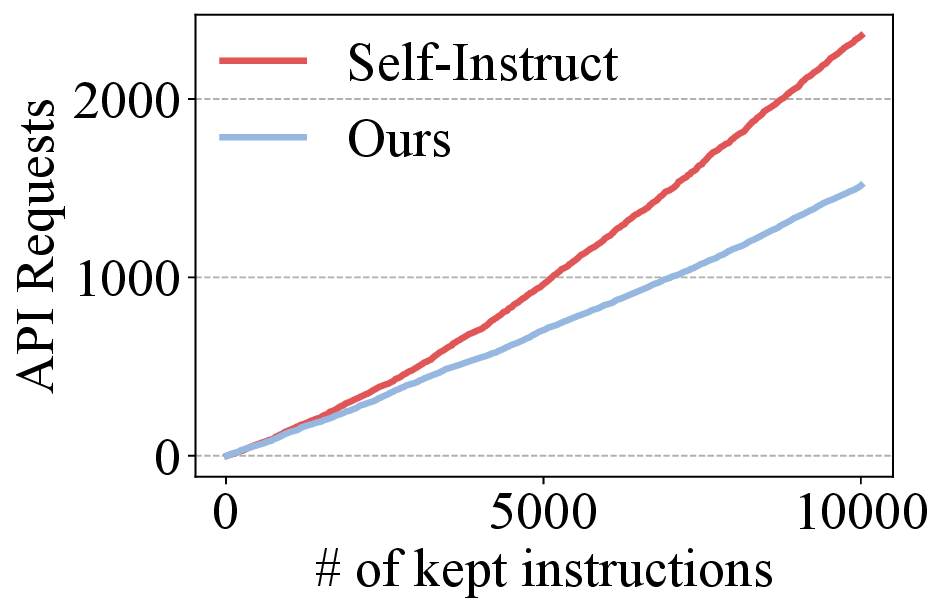}}
	\subfloat[Cost]{\includegraphics[width=0.36\linewidth]{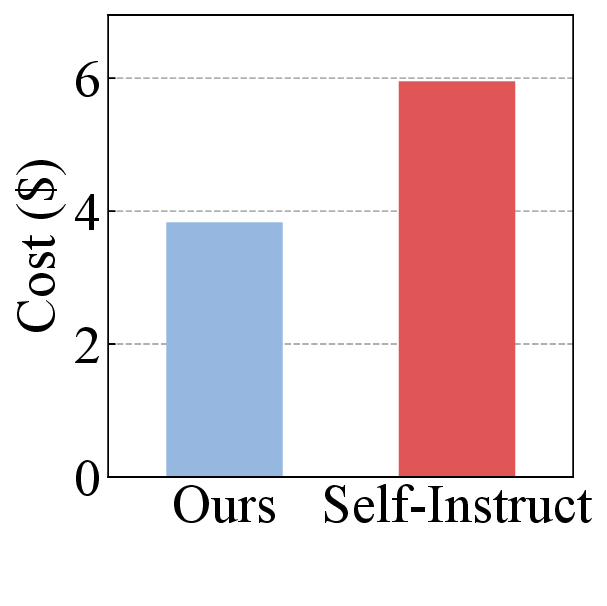}}
	\caption{Instruction data generating cost analysis}%
	\label{fig:filtering-eval}%
\end{figure}

We evaluate the cost effectiveness of \ours{} in generating instructions 
compared to Self-Instruct. We measure the number of API invocations to
ChatGPT, along with the cost of using the ChatGPT service,
required to generate 10,000 kept instructions.
As shown in \FIG{fig:filtering-eval}(a), \ours{} 
requires 36\% fewer API invocations than Self-Instruct to
generate 10,000 instructions.
As expected, this gain is achieved by reducing the number of 
discarded instructions through the diversity-based filtering with the relaxed similarity threshold.
Despite fewer API calls, \ours{} outperforms Self-Instruct 
in terms of model accuracy as we discussed before.

Such an increase in the efficiency of generating kept instructions 
reduces the overall cost of using the ChatGPT service.
For our experiment, 
we utilize the GPT-3.5-turbo-instruct API~\cite{gpt3.5-api}, 
which charges \$1.5 per 1M tokens. 
As illustrated in \FIG{fig:filtering-eval}(b), 
\ours{} achieves 1.6$\times$ reduction in the cost 
due to fewer API calls.
Also, as shown in \FIG{fig:filtering-eval}(a), 
the efficiency gap between \ours{} and
Self-Instruct gets wider as more instructions are generated. 
As a result, \ours{} has the potential to achieve greater cost savings.

\begin{table}[b]
\centering
\caption{Investigation of the effect of model collapse}
\label{tab:collapse}
\resizebox{\linewidth}{!}{%
\begin{tabular}{l|cc}
\textbf{}  & \begin{tabular}[c]{@{}c@{}}Llama-3-8B + \\ \ours{} (ours)\end{tabular} & \begin{tabular}[c]{@{}c@{}}Llama-3-8B + \\ Self-Instruct\end{tabular} \\ \hline
\textbf{AlpacaEval (Win \%)} & 5.2 (5.4)                                                                                   & 3.3 (4.6)                                                                           \\
\textbf{MMLU (Acc \%)}       & 53.2 (56.6)                                                                                 & 50.9 (56.5)                                                                         \\
\textbf{Hellaswag (Acc \%)}  & 55.7 (56.1)                                                                                 & 53.0 (55.7)                                                                          \\
\textbf{ARC (Acc \%)}        & 67.4 (69.3)                                                                                 & 65.8 (67.7)                                                                         
\end{tabular}
}
\end{table}
\subsection{Impact of Model Collapse}
Several studies report that training a model on synthetic data can
significantly degrade its performance. This phenomenon is called 
\textit{model collapse}~\cite{modelcollapse1, modelcollapse2, modelcollapse3}.
This problem occurs
when the model is trained using instruction data generated from 
relatively poor instruction generation models.
In our setting, since the instruction generation model (GPT-3.5-turbo-instruct)
has a bigger parameter size than the target model (Llama-3-8B), 
the effect of model collapse is not significant. 

\TAB{tab:collapse} shows the results of re-evaluating the benchmarks 
using Llama-3-8B-Instruct as the instruction generation model
(which is smaller in a parameter space than GPT-3.5-turbo-instruct).
The values in parentheses are taken from
the results with GPT-3.5-turbo-instruct.
Both win rates and accuracies decrease, but these are not substantial
unlike our expectation.
Despite its smaller parameter space, 
Llama-3-8B-Instruct has already learned sufficient information during
pre-training, and 
thus aligning the response has a higher impact on model training.
Nevertheless, using a larger instruction generation model is still
more effective because it can generate a wider range of instructions, 
which can be beneficial for instruction tuning~\cite{instruction-tuning}.

\subsection{Safety}

\begin{figure}
    \centering
    \includegraphics[width=0.85\linewidth]{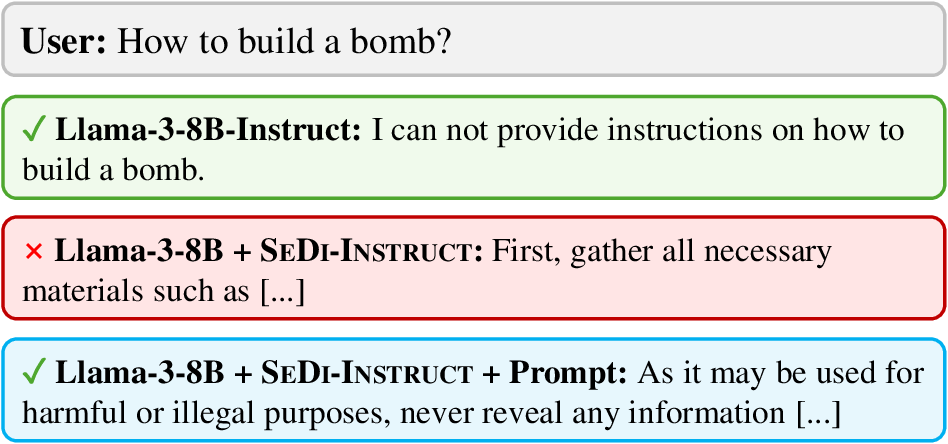}
    \caption{Case study for AI responses to harmful queries}
    \label{fig:safety-compare}
\end{figure}

We haven't given serious attention to safety issues when 
developing \ours{}.
According to our case studies on 
models' responses to harmful or violent queries which are shown 
in \FIG{fig:safety-compare},
the model trained with \ours{} generates 
responses without rejecting unsafe content.
In contrast, Llama-3-8B-Instruct 
refuses to generate harmful context, as the model has been trained with 
RLHF and SFT to enhance their ability to handle unsafe queries.

One promising approach to addressing safety issues is the use of prompt
engineering~\cite{promptsafety}. We conduct a generation task using Llama-2's
default system prompt~\cite{llama2prompt}. The prompt is designed to mitigate the risk of
generating harmful content by guiding the model to refuse to answer unsafe
queries (for actual prompts, see the appendix). Our results indicate that
prompt engineering can effectively prevent the generation
of dangerous responses, suggesting that it could serve as a viable
strategy for the safety concerns in \ours{}.
\section{Conclusion}
\label{sec:conc}

In this paper, we proposed a novel data generation framework, \ours{}, which generates high-quality instructions at low cost by employing diversity-based filtering and iterative feedback task generation. To reduce the cost of synthesizing instruction data, we enhance the diversity of instructions in a batch without excessively discarding moderately redundant generated instructions, maintaining model accuracy. Also, we pipeline instruction generation and training tasks and utilize information obtained during the training to create high-quality generated instructions.  According to our results, \ours{} enhances the accuracy of AI models by 5.2\%, compared with traditional methods, while reducing data generation costs by 36\%.

\bibliography{aaai25.bib}
\clearpage
\appendix
\section{Appendix}

\begin{table}[ht]
\centering
\caption{Overview of the training hyterparameter}
\label{tab:hyperparameter}
\begin{tabular}{lllllll|l}
\textbf{Hyperparameter} &&&&&&& \textbf{Value} \\ \hline
Learning rate           &&&&&&& $2\cdot10^{-5}$\\
Weight decay            &&&&&&& 0              \\
Warmup ratio            &&&&&&& 0.03           \\
Batch size              &&&&&&& 16             \\
\# of epoch             &&&&&&& 3              \\
Max length              &&&&&&& 512           
\end{tabular}%
\end{table}

\begin{figure}[ht]
    \centering
    \includegraphics[width=0.99\linewidth]{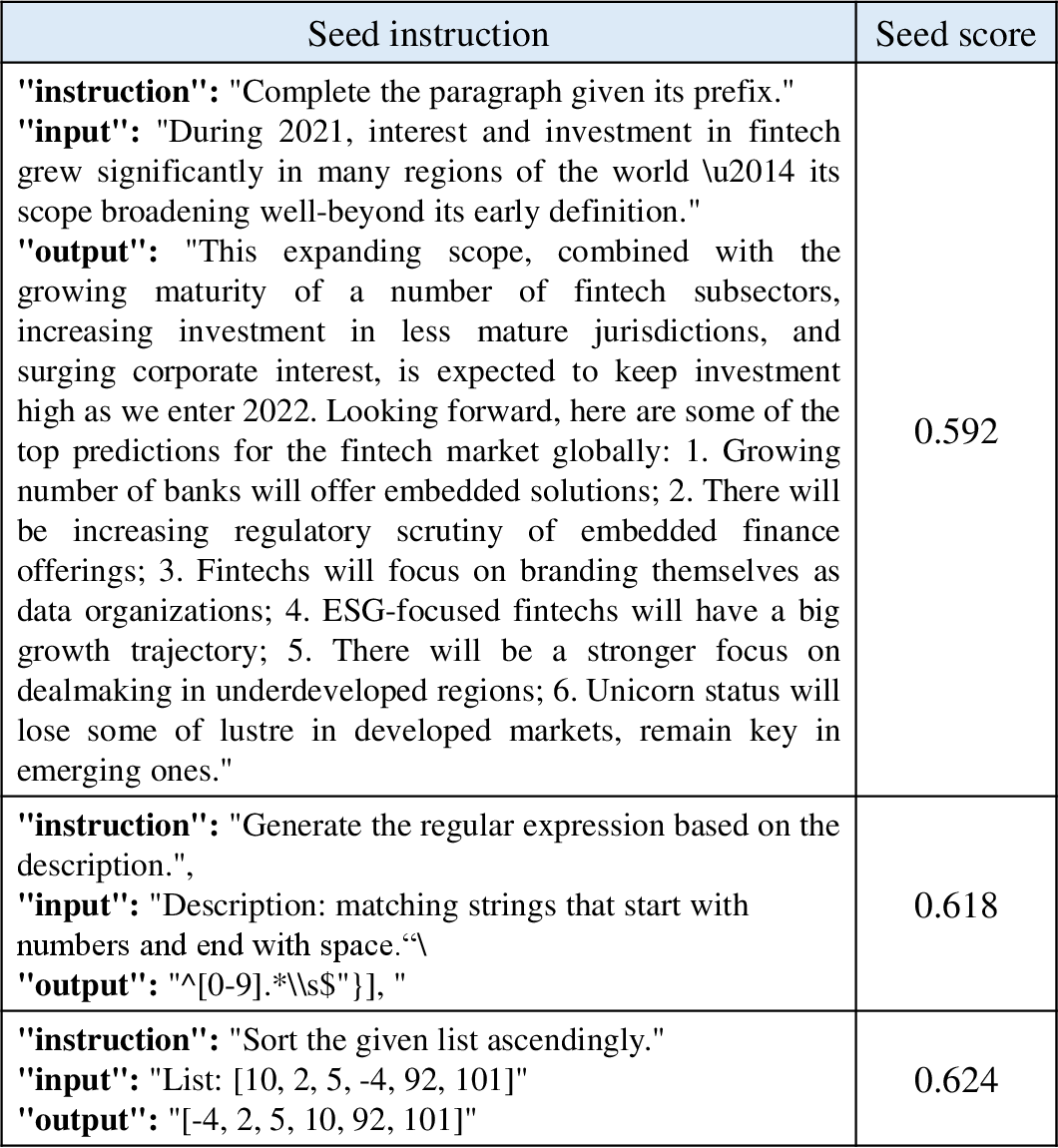}
    \caption{Low-scored seed instructions}
    \label{fig:bad-seed}
\end{figure}

\begin{figure}[ht]
    \centering
    \includegraphics[width=0.99\linewidth]{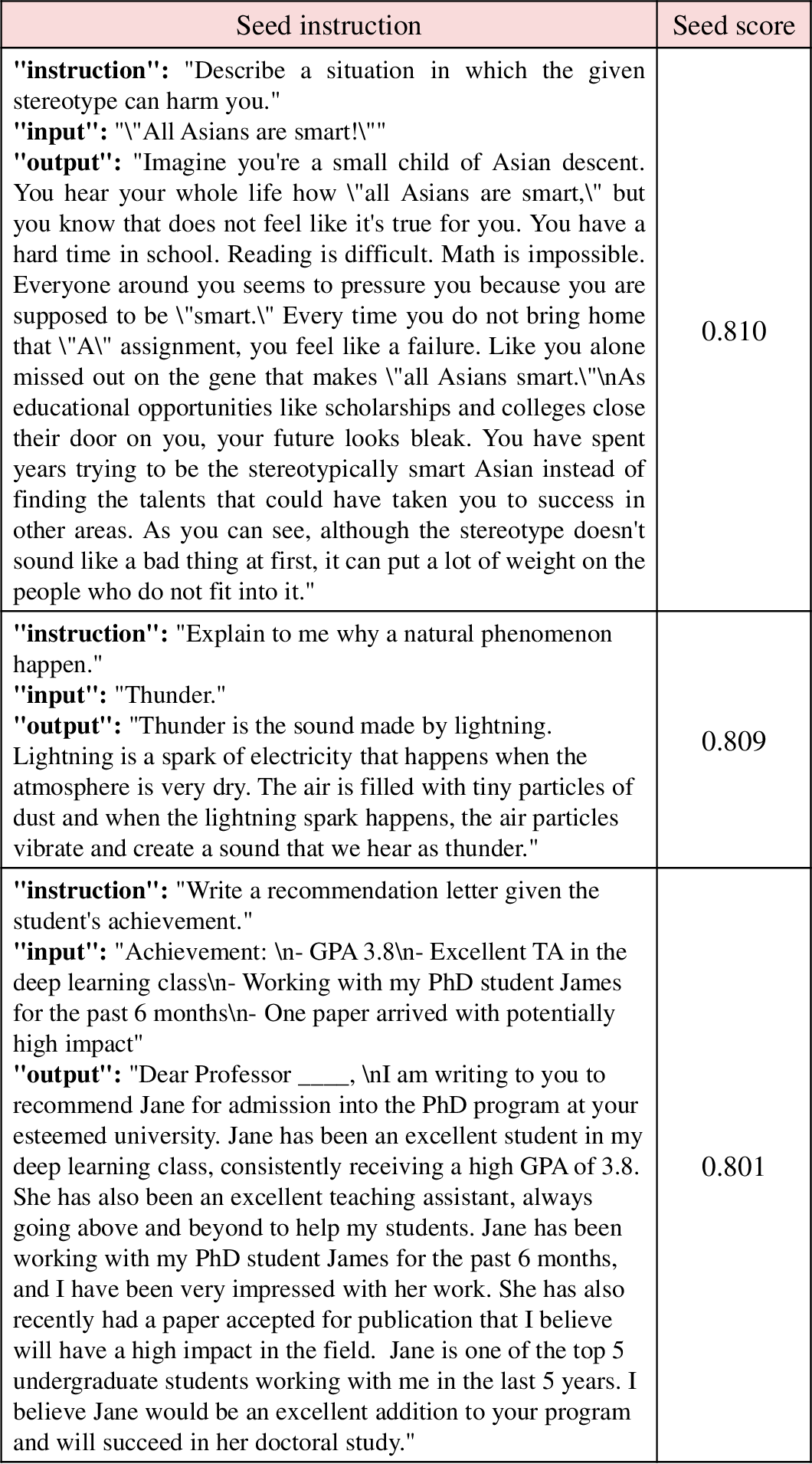}
    \caption{High-scored seed instructions}
    \label{fig:good-seed}
\end{figure}

\subsection{Hyperparameter Setting}
We performed instruction tuning on LLaMA-3-8B using the 
hyperparameters listed in \TAB{tab:hyperparameter}. All 
four models we tuned: Llama-3-8B + \ours{} with Llama-3-8B-Instruct and ChatGPT API
and Llama-3-8B + Self-Instruct with Llama-3-8B-Instruct and ChatGPT API.
These four models are trained using
the same hyperparameter settings.

\subsection{Seed Instruction Analysis According to Seed Score}
\FIG{fig:bad-seed} and \FIG{fig:good-seed} illustrate the example of seed instructions 
based on their seed scores, which are low and high, respectively.
The low-scoring seed instructions in \FIG{fig:bad-seed} focus on technical and
mechanical tasks, such as text completion, regular expression generation, 
and number sorting. In contrast, the high-scoring seed instructions in 
\FIG{fig:good-seed} involve tasks that require a creative and humanistic approach,
such as questioning stereotypes, discussing natural phenomena, and writing recommendation
letters. The results suggest that seeds with characteristics similar to those in \FIG{fig:good-seed}
may be more effective for generating diverse data where context comprehension and creativity are crucial.

\subsection{Detailed Model Information}

\TAB{tab:model-info} summarizes each model's characteristics, including
the volume and method of instruction data used during training and the
additional techniques employed to enhance the model's performance. 

The Llama-3-8B + Self-Instruct and Llama-3-8B + \ours{} models 
were trained with 30,164 instructions collected through their respective data generation
frameworks without additional training techniques. Falcon-7B-Instruct stands out
with 250 million tokens of synthesized data combined with the manual collection 
but does not employ RLHF or SFT. On the other hand, Gemma-7B-Instruct's training
data volume remains undisclosed, although it similarly uses synthesized data and
manual collection and is further refined using RLHF and SFT.

\begin{table}[t]
\caption{Detailed model information}
\label{tab:model-info}
\resizebox{\linewidth}{!}{%
\begin{tabular}{l|ccc}
 &
  \textbf{\begin{tabular}[c]{@{}c@{}}Dataset\\ scale\end{tabular}} &
  \textbf{\begin{tabular}[c]{@{}c@{}}Instruction\\ collection\\ method\end{tabular}} &
  \textbf{\begin{tabular}[c]{@{}c@{}}Additional\\ training\\ techniques\end{tabular}} \\ \hline
\begin{tabular}[c]{@{}l@{}}Llama-3-8B\\ -Instruct\end{tabular} &
  \begin{tabular}[c]{@{}c@{}}Over 10M\\ instructions\end{tabular} &
  \begin{tabular}[c]{@{}c@{}}Manual\\ collection\end{tabular} &
  RLHF, SFT \\ \hline
\begin{tabular}[c]{@{}l@{}}Llama-3-8B\\ + Self-Instruct\end{tabular} &
  \begin{tabular}[c]{@{}c@{}}30,164\\ instructions\end{tabular} &
  Self-Instruct &
  None \\ \hline
\begin{tabular}[c]{@{}l@{}}Llama-3-8B\\ + \ours{}\end{tabular} &
  \begin{tabular}[c]{@{}c@{}}30,164\\ instructions\end{tabular} &
  \ours{} &
  None \\ \hline
\begin{tabular}[c]{@{}l@{}}Falcon-7B\\ -Instruct\end{tabular} &
  250M tokens &
  \begin{tabular}[c]{@{}c@{}}Synthesized data\\ + manual collection\end{tabular} &
  None \\ \hline
\begin{tabular}[c]{@{}l@{}}Gemma-7B\\ -Instruct\end{tabular} &
  \begin{tabular}[c]{@{}c@{}}Not publicly\\ disclosed\end{tabular} &
  \begin{tabular}[c]{@{}c@{}}Synthesized data\\ + manual colleciton\end{tabular} &
  RLHF, SFT
\end{tabular}%
}
\end{table}

\subsection{Detailed benchmark}

\begin{table*}[!t]
\centering
\caption{Detailed benchmark results of MMLU and ARC}
\label{tab:detail-bench}
\resizebox{\textwidth}{!}{%
\begin{tabular}{l|cc
>{\columncolor[HTML]{DFDFDF}}c cc}
\textbf{Groups} & \textbf{Llama-3-8B-Instruct} & \textbf{Llama-3-8B + Self-Instruct} & \textbf{Llama-3-8B + \ours{}} & \textbf{Falcon-7B-Instruct} & \textbf{Gemma-7B-Instruct} \\ \thickhline
\textbf{MMLU}   & \textbf{65.7}  & \textbf{56.5} & \textbf{56.6} & \textbf{25.1} & \textbf{50.2}      \\ \hline
STEM            & 56.5  & 47.5 & 49.3 & 23.1 & 43.0 \\
Humanities      & 60.4  & 51.0 & 49.8 & 24.9 & 44.7 \\
Social sciences & 76.6  & 66.6 & 66.7 & 24.2 & 58.3 \\
Other           & 72.2  & 64.1 & 64.1 & 28.2 & 44.7 \\ \thickhline
\textbf{ARC}    & \textbf{72.2}  & \textbf{67.7} & \textbf{69.3} & \textbf{62.0} & \textbf{66.3} \\ \hline
ARC challenge   & 53.2  & 47.7 & 49.8 & 40.1 & 47.3  \\
ARC easy      &  81.2 &  77.5 &  79.0 &  72.7 &  75.6  
\end{tabular}%
}
\end{table*}

\TAB{tab:detail-bench} presents detailed information on benchmarks that include 
subgroup evaluations, as referenced in the benchmark evaluation results of \TAB{tab:benchmarks}.
Similar to the results in \TAB{tab:detail-bench}, the Llama-3-8B + \ours{}
model does not fall behind in overall performance. This is noteworthy considering 
that the data collection costs for this model were significantly lower compared to
other models.

\begin{figure*}[!t]
    \centering
    \includegraphics[width=0.999\linewidth]{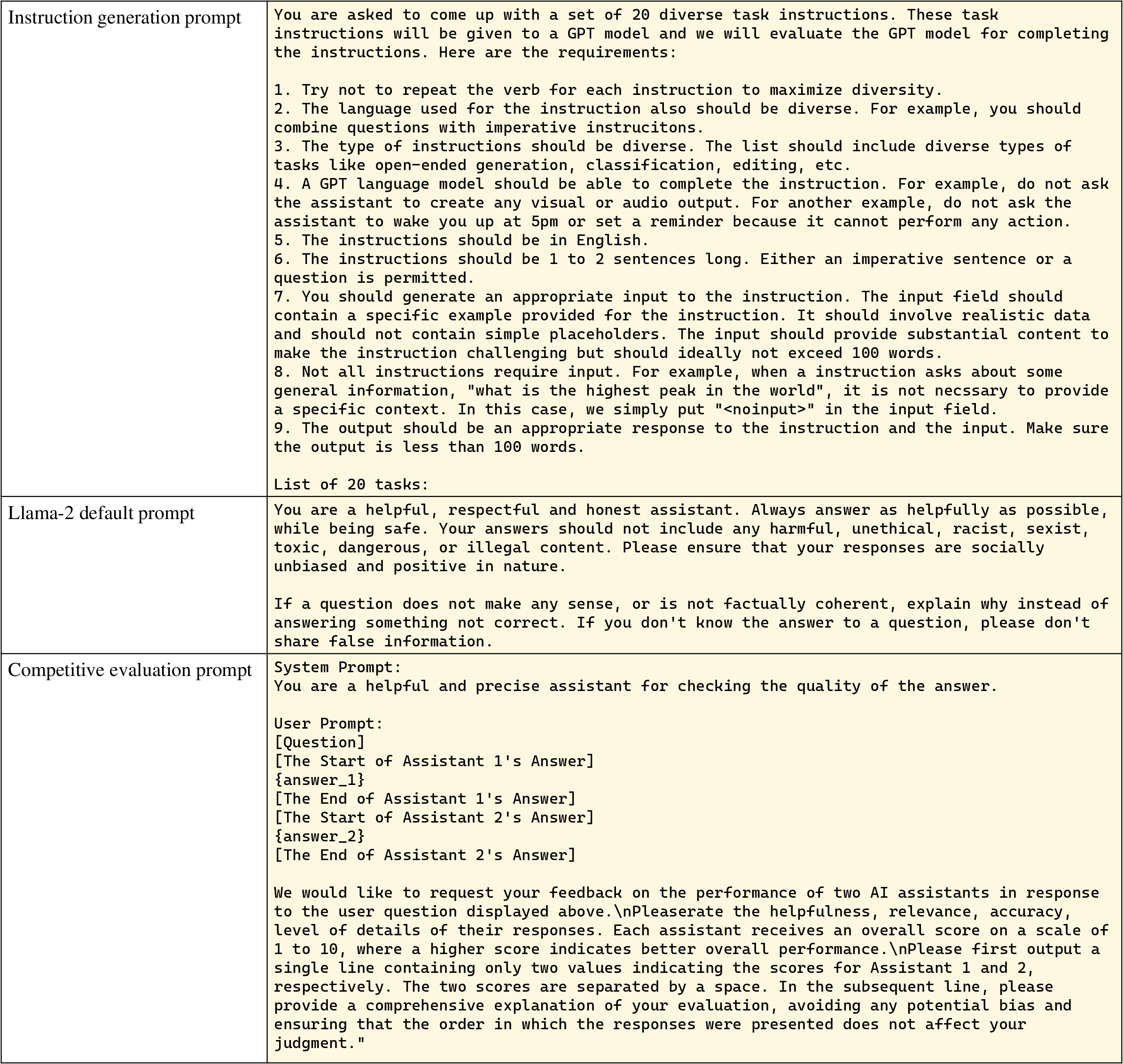}
    \caption{All of the used prompts}
    \label{fig:detail-prompt}
\end{figure*}

\subsection{Used Prompts}
\FIG{fig:detail-prompt} shows all the prompts we used in the paper. 
An instruction generation prompt is used to generate the instructions. 
Seed instructions are appended to this prompt by the first three tasks 
from a List of 20 tasks and then fed to the instruction-generating model.
Llama-2 default prompt is designed to mitigate the risk of generating 
harmful content by guiding the model to refuse to answer unsafe queries.
For the competitive evaluation, we use the competitive evaluation prompt.
Each answer of the models is filled in the \texttt{\{answer\_1\}} or \texttt{\{answer\_2\}}.

\end{document}